\documentclass[final]{cvpr}

\usepackage{times}
\usepackage{epsfig}
\usepackage{graphicx}
\usepackage{amsmath}
\usepackage{amssymb}
\usepackage{gensymb}

\usepackage[pagebackref=true,breaklinks=true,colorlinks,bookmarks=false]{hyperref}

\usepackage[usenames, dvipsnames]{color} 
\definecolor{purple}{rgb}{0.5, 0.0, 0.5}
\definecolor{orange}{rgb}{1, 0.65, 0}
\definecolor{lightgreen}{rgb}{0.68, 1, 0.18}
\definecolor{darkgreen}{rgb}{0.09, 0.32, 0.24}
\definecolor{darkred}{rgb}{0.6, 0, 0}
\definecolor{brown}{rgb}{0.64, 0.16, 0.16}
\definecolor{lilac}{rgb}{0.82, 0.67, 1}
\definecolor{purple}{rgb}{0.5, 0.0, 0.5}
\definecolor{cerulean}{rgb}{0.0, 0.43, 0.65}

\newcommand{\mypar}[1]{\vspace{-2mm}\paragraph{#1}} 
\usepackage{enumitem}

\def\cvprPaperID{13} 
\def\confYear{CVPR 2021 ADP3 workshop}

\begin{document}

\title{nuPlan: A closed-loop ML-based planning benchmark for autonomous vehicles}

\author{
Holger Caesar\;\;\;
Juraj Kabzan\;\;\;
Kok Seang Tan\;\;\;
Whye Kit Fong\;\;\;
Eric Wolff\\
Alex Lang\;\;\;
Luke Fletcher\;\;\;
Oscar Beijbom\;\;\;
Sammy Omari\\
Motional
}

\maketitle

\begin{abstract}
\vspace{-2mm}
In this work, we propose the world's first closed-loop ML-based planning benchmark for autonomous driving.
While there is a growing body of ML-based motion planners, the lack of established datasets and metrics has limited the progress in this area. 
Existing benchmarks for autonomous vehicle motion prediction have focused on short-term motion forecasting, rather than long-term planning.
This has led previous works to use open-loop evaluation with L2-based metrics, which are not suitable for fairly evaluating long-term planning.
Our benchmark overcomes these limitations by introducing a large-scale driving dataset, lightweight closed-loop simulator, and motion-planning-specific metrics.
We provide a high-quality dataset with 1500h of human driving data from 4 cities across the US and Asia with widely varying traffic patterns (Boston, Pittsburgh, Las Vegas and Singapore).
We will provide a closed-loop simulation framework with reactive agents and provide a large set of both general and scenario-specific planning metrics.
We plan to release the dataset at NeurIPS 2021 and organize benchmark challenges starting in early 2022.
\end{abstract}

\vspace{-4mm}
\section{Introduction}
\label{sec:introduction}
Large-scale human labeled datasets in combination with deep Convolutional Neural Networks have led to an impressive performance increase in autonomous vehicle~(AV) perception over the last few years~\cite{kitti,nuscenes}.
In contrast, existing solutions for AV planning are still primarily based on carefully engineered expert systems, that require significant amounts of engineering to adapt to new geographies and do not scale with more training data.
We believe that providing suitable data and metrics will enable ML-based planning and pave the way towards a full ``Software 2.0'' stack.

\begin{figure}[t]
\begin{center}
\vspace{-2mm}
\includegraphics[width=\linewidth, height=1.5cm]{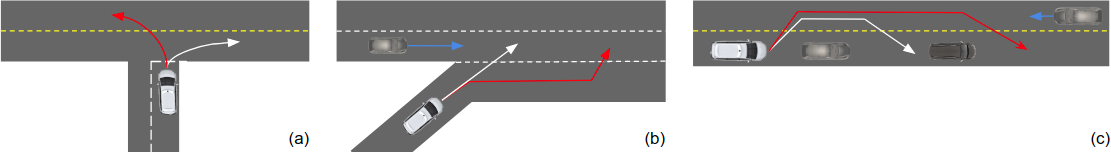}
\end{center}
\vspace{-2mm}
\caption{
We show different driving scenarios to emphasize the limitations of existing benchmarks.
The observed driving route of the ego vehicle in shown in white and the hypothetical planner route in red. 
(a) The absence of a goal leads to ambiguity at intersections.
(b) Displacement metrics do not take into account the multi-modal nature of driving.
(c) open-loop evaluation does not take into account agent interaction.
}
\vspace{-4mm}
\label{fig:main}
\end{figure}

Existing real-world benchmarks are focused on short-term motion forecasting, also known as prediction~\cite{argoverse,nuscenes,lyft,waymo}, rather than planning.
This is evident in the lack of high-level goals, the choice of metrics, and the open-loop evaluation.
Prediction focuses on the behavior of other agents, while planning relates to the ego vehicle behavior.
Prediction is typically multi-modal, which means that for each agent we predict the $N$ most likely trajectories.
In contrast, planning is typically uni-modal (except for contingency planning) and we predict a single trajectory.
As an example, in Fig.~\ref{fig:main}a, turning left or right at an intersection are equally likely options.
Prediction datasets lack a baseline navigation route to indicate the high-level goals of the agents.
In Fig.~\ref{fig:main}b, the options of merging immediately or later are both equally valid, but the commonly used L2 distance-based metrics (minADE, minFDE, and miss rate) penalize the option that was not observed in the data.
Intuitively, the distance between the predicted trajectory and the observed trajectory is not a suitable indicator in a multi-modal scenario. 
In Fig.~\ref{fig:main}c, the decision whether to continue to overtake or get back into the lane should be based on the consecutive actions of all agent vehicles, which is not possible in open-loop evaluation.
Lack of closed-loop evaluation leads to systematic drift, making it difficult to evaluate beyond a short time horizon (3-8s). 

We instead provide a planning benchmark to address these shortcomings. Our main contributions are:
\begin{itemize}[leftmargin=*]
\itemsep0em 
\vspace{-2mm}
    \item The largest existing public real-world dataset for autonomous driving with high quality autolabeled tracks from 4 cities.
    \item Planning metrics related to traffic rule violation, human driving similarity, vehicle dynamics, goal achievement, as well as scenario-based.
    \item The first public benchmark for real-world data with a closed-loop planner evaluation protocol.
\end{itemize}

\vspace{-2mm}
\section{Related Work}
We review the relevant literature for prediction and planning datasets, simulation, and ML-based planning.

\mypar{Prediction datasets.}
Table~\ref{tab:datasets} shows a comparison between our dataset and relevant prediction datasets.
Argoverse Motion Forecasting~\cite{argoverse} was the first large-scale prediction dataset.
With 320h of driving data, it was unprecedented in size and provides simple semantic maps with centerlines and driveable area annotations.
However, the autolabeled trajectories in the dataset are of lower quality due to the state of object detection field at the time and the insufficient amount of human-labeled training data (113 scenes).

The nuScenes prediction~\cite{nuscenes} challenge consists of 850 human-labeled scenes from the nuScenes dataset.
While the annotations are high quality and sensor data is provided, the small scale limits the number of driving variations.
The Lyft Level 5 Prediction Dataset~\cite{lyft} contains 1118h of data from a single route of 6.8 miles.
It features detailed semantic maps, aerial maps, and dynamic traffic light status.
While the scale is unprecedented, the autolabeled tracks are often noisy and geographic diversity is limited.
The Waymo Open Motion Dataset~\cite{waymo} focuses specifically on the interactions between agents, but does so using open-loop evaluation.
While the dataset size is smaller than existing datasets at 570h, the autolabeled tracks are of high quality~\cite{waymo_offboard}.
They provide semantic maps and dynamic traffic light status.

These datasets focus on prediction, rather than planning.
In this work we aim to overcome this limitation by using planning metrics and closed-loop evaluation.
We are the first large-scale dataset to provide sensor data.

\begin{table}[t]
    \centering
    \resizebox{\linewidth}{!}{
    \begin{tabular}{c|c c c c c c}
         Dataset        &   Data    &   Cities  &  Sensor Data     &   Type       &   Evaluation  \\
         \hline
         Argoverse      &   320h    &   2       &                  &   Pred       &   OL         \\
         nuPredict      &   5h      &   2       &      \checkmark  &   Pred       &   OL         \\
         Lyft           &   1118h   &    1      &                  &   Pred       &   OL         \\
         Waymo          &   570h    &   6       &                  &   Pred       &   OL         \\
         nuPlan         & 1500h     &   4       &     \checkmark   &   Plan.      &   OL+CL      \\
    \end{tabular}
    }
    \vspace{-2mm}
    \caption{A comparison of leading datasets for motion prediction (Pred) and planning (Plan). We show the dataset size, number of cities, availability of sensor data, dataset type, and whether it uses open-loop (OL) or closed-loop (CL) evaluation. nuPredict refers to the prediction challenge of the nuScenes~\cite{nuscenes} dataset.
    }
    \vspace{-2mm}
    \label{tab:datasets}
\end{table}

\mypar{Planning datasets.}
CommonRoad~\cite{commonroad} provides a first of its kind planning benchmark, that is composed of different vehicle models, cost functions and scenarios (including goals and constraints).
There are both pre-recorded and interactive scenarios.
With 5700 scenarios in total, the scale of the dataset does not support training modern deep learning based methods.
All scenarios lack sensor data.

\mypar{Simulation.}
Simulators have enabled breakthroughs in planning and reinforcement learning with their ability to simulate physics, agents, and environmental conditions in a closed-loop environment.

AirSim~\cite{airsim2017fsr} is a high-fidelty simulator for AVs, such as drones and cars.
It includes a physics engine that can operate at a high frequency for real-time hardware-in-the-loop simulation.
CARLA~\cite{carla} supports the training and validation of autonomous urban driving systems.
It allows for flexible specification of sensor suites and environmental conditions. 
In the CARLA Autonomous Driving Challenge\footnote{See \url{carlachallenge.org}} the goal is to navigate a set of waypoints using different combinations of sensor data and HD maps. Alternatively, users can use scene abstraction to omit the perception task and focus on planning and control aspects. This challenge is conceptually similar to what we propose, but does not use real world data and provides less detailed planning metrics.

Sim-to-real transfer is an active research area for diverse tasks such as localization, perception, prediction, planning and control.
\cite{track_with_permanence} show that the domain gap between simulated and real-world data remains an issue, by transferring a synthetically trained tracking model to the KITTI~\cite{kitti} dataset. To overcome the domain gap, they jointly train their model using real-world data for visible and simulation data for occluded objects.
\cite{learning_to_drive} learn how to drive by transferring a vision-based lane following driving policy from simulation to the real world without any real-world labels.
\cite{simulation_reinforcement_learning} use reinforcement learning in simulation to obtain a driving system controlling a full-size real-world vehicle. They use mostly synthetic data, with labelled real-world data appearing only in the training of the segmentation network.

However, all simulations have fundamental limits since they introduce systematic biases. More work is required to plausibly emulate real-world sensors, e.g. to generate photo-realistic camera images.

\mypar{ML-based planning.}
A new emerging research field is ML-based planning for AVs using real-world data.
However, the field has yet to converge on a common input/output space, dataset, or metrics.
A jointly learnable behavior and trajectory planner is proposed in~\cite{joint_behavior_trajectory_planning}.
An interpretable cost function is learned on top of models for perception, prediction and vehicle dynamics, and evaluated in open-loop on two unpublished datasets.
An end-to-end interpretable neural motion planner~\cite{e2e_inmp} takes raw lidar point clouds and dynamic map data as inputs and predicts a cost map for planning. 
They evaluate in open-loop on an unpublished dataset, with a planning horizon of only 3s.
ChauffeurNet~\cite{chauffeurnet} finds that standard behavior cloning is insufficient for handling complex driving scenarios, even when using as many as 30 million examples.
They propose exposing the learner to synthesized data in the form of perturbations to the expert’s driving and augment the imitation loss with additional losses that penalize undesirable events and encourage progress. 
Their unpublished dataset contains 26 million examples which correspond to 60 days of continuous driving.
The method is evaluated in a closed-loop and an open-loop setup, as well as in the real world.
They also show that open-loop evaluation can be misleading compared to closed-loop.
MP3~\cite{mp3} proposes an end-to-end approach to mapless driving, where the input is raw lidar data and a high-level navigation goal.
They evaluate on an unpublished dataset in open and closed-loop.
Multi-modal methods have also been explored in recent works \cite{transfuser,cil_e2e_mmsfs,cil_cam_lidar_fusion}.
These approaches explore different strategies for fusing various modality representations in order to predict future waypoints or control commands.
Neural planners were also used in \cite{pkl,nuscenespkl} to evaluate an object detector using the KL divergence of the planned trajectory and the observed route.

Existing works evaluate on different metrics which are inconsistent across the literature.
TransFuser~\cite{transfuser} evaluates its method on the number of infractions, the percentage of the route distance completed, and the route completion weighted by an infraction multiplier.
Infractions include collisions with other agents, and running red lights.
\cite{cil_e2e_mmsfs} evaluates its planner using off-road time, off-lane time and number of crashes, while \cite{cil_cam_lidar_fusion,cil_multimodal_e2e} report the success rate of reaching a given destination within a fixed time window.
\cite{cil_cam_lidar_fusion} also introduces another metric which measures the average percentage of distance travelled to the goal.

While ML-based planning has been studied in great detail, the lack of published datasets and a standard set of metrics that provide a common framework for closed-loop evaluation has limited the progress in this area.
We aim to fill this gap by providing an ML-based planning dataset and metrics. 

\section{Dataset}
\label{sec:dataset}

\mypar{Overview.}
We plan to release 1500 hours of data from Las Vegas, Boston, Pittsburgh, and Singapore.
Each city provides its unique driving challenges.
For example, Las Vegas includes bustling casino pick-up and drop-off points (PUDOs) with complex interactions and busy intersections with up to 8 parallel driving lanes per direction, Boston routes include drivers who love to double park, Pittsburgh has its own custom precedence pattern for left turns at intersections, and Singapore features left hand traffic.
For each city we provide semantic maps and an API for efficient map queries.
The dataset includes lidar point clouds, camera images, localization information and steering inputs.
While we release autolabeled agent trajectories on the entire dataset, we make only a subset of the sensor data available due to the vast scale of the dataset (200+ TB).

\mypar{Autolabeling.}
We use an offline perception system to label the large-scale dataset at high accuracy, without the real-time constraints imposed on the online perception system of an AV.
We use PointPillars~\cite{pointpillars} with CenterPoint~\cite{centerpoint}, a modified version multi-view fusion (MVF++)~\cite{waymo_offboard}, and non-causal tracking to achieve near-human labeling performance. 

\mypar{Scenarios.}
To enable scenario-based metrics, we automatically annotate intervals with tags for complex scenarios.
These scenarios include merges, lane changes, protected or unprotected left or right turns, interaction with cyclists, interaction with pedestrians at crosswalks or elsewhere, interactions with close proximity or high acceleration, double parked vehicles, stop controlled intersections and driving in construction zones.

\section{Benchmarks}
To further the state of the art in ML-based planning, we organize benchmark challenges with the tasks and metrics described below.

\subsection{Overview}
To evaluate a proposed method against the benchmark dataset, users submit ML-based planning code to our evaluation server.
The code must follow a provided template. Contrary to most benchmarks, the code is containerized for portability in order to enable closed-loop evaluation on a secret test set.
The planner operates either on the autolabeled trajectories or, for end-to-end open-loop approaches, directly on the raw sensor data.
When queried for a particular timestep, the planner returns the \emph{planned} position and heading of the ego vehicle.
A provided controller will then drive a vehicle while closely tracking the planned trajectory.
We use a predefined motion model to simulate the ego vehicle motion in order to approximate a real system.
The final driven trajectory is then scored against the metrics defined in Sec~\ref{sec:metrics}.

\subsection{Tasks}
We present the three different tasks for our dataset with increasing difficulty. 

\mypar{Open-loop.}
In the first challenge, we task the planning system to mimic a human driver.
For every timestep, the trajectory is scored based on predefined metrics.
It is not used to control the vehicle.
In this case, no interactions are considered. 

\mypar{Closed-loop.}
In the closed-loop setup the planner outputs a \emph{planned} trajectory using the information available at each timestep, similar to the previous case.
However, the proposed trajectory is used as a reference for a controller, and thus, the planning system is gradually corrected at each timestep with the new state of the vehicle.
While the new state of the vehicle may not coincide with that of the recorded state, leading to different camera views or lidar point clouds, we will not perform any sensor data warping or novel view synthesis.
In this set, we distinguish between two tasks.
In the \emph{Non-reactive closed-loop} task we do not make any assumptions on other agents behavior and simply use the observed agent trajectories.
As shown in~\cite{lyft}, the vast majority of interventions in closed-loop simulation is due to the non-reactive nature, e.g. vehicles naively colliding with the ego vehicle.
In the \emph{reactive closed-loop} task we provide a planning model for all other agents that are tracked like the ego vehicle.

\label{sec:metrics}
\subsection{Metrics}
We split the metrics into two categories, common metrics, which are computed for every scenario and scenario-based metrics, which are tailored to predefined scenarios. 

\mypar{Common metrics.}
\begin{itemize}[leftmargin=*]
    \item \emph{Traffic rule violation} is used to measure compliance with common traffic rules. We compute the rate of collisions with other agents, rate of off-road trajectories, the time gap to lead agents, time to collision and the relative velocity while passing an agents as a function of the passing distance. 
    \item \emph{Human driving similarity} is used to quantify a maneuver satisfaction in comparison to a human, e.g. longitudinal velocity error, longitudinal stop position error and lateral position error. In addition, the resulting jerk/acceleration is compared to the human-level jerk/acceleration.
    \item \emph{Vehicle dynamics} quantify rider comfort and feasibility of a trajectory. Rider comfort is measured by jerk, acceleration, steering rate and vehicle oscillation. Feasibility is measured by violation of predefined limits of the same criteria. 
    \item \emph{Goal achievement} measures the route progress towards a goal waypoint on the map using L2 distance.
\end{itemize}

\mypar{Scenario-based metrics.}
Based on the scenario tags from Sec.~\ref{sec:dataset}, we use additional metrics for challenging maneuvers.
For \emph{lane change}, time to collision and time gap to lead/rear agent on the target lane is measured and scored.
For \emph{pedestrian/cyclist interaction}, we quantify the passing relative velocity while differentiating their location.
Furthermore, we compare the \emph{agreement between decisions made by a planner and human} for crosswalks and unprotected turns (right of way). 

\mypar{Community feedback.}
Note that the metrics shown here are an initial proposal and do not form an exhaustive list.
We will work closely with the community to add novel scenarios and metrics to achieve consensus across the community. 
Likewise, for the main challenge metric we see multiple options, such as a weighted sum of metrics, a weighted sum of metric violations above a predefined threshold or a hierarchy of metrics.
We invite the community to collaborate with us to define the metrics that will drive this field forward.

\section{Conclusion}
In this work we proposed the first ML-based planning benchmark for AVs.
Contrary to existing forecasting benchmarks, we focus on goal-based planning, planning metrics and closed-loop evaluation.
We hope that by providing a common benchmark, we will pave a path towards progress in ML-based planning, which is one of the final frontiers in autonomous driving.

{\small
\bibliographystyle{ieee_fullname}
\bibliography{egbib}

\begin{thebibliography}{10}\itemsep=-1pt

\bibitem{commonroad}
Matthias Althoff, Markus Koschi, and Stefanie Manzinger.
\newblock {CommonRoad}: Composable benchmarks for motion planning on roads.
\newblock In {\em Proc. of the IEEE Intelligent Vehicles Symposium}, 2017.

\bibitem{chauffeurnet}
Mayank Bansal, Alex Krizhevsky, and Abhijit Ogale.
\newblock Chauffeurnet: Learning to drive by imitating the best and
  synthesizing the worst.
\newblock In {\em RSS}, 2019.

\bibitem{learning_to_drive}
Alex Bewley, Jessica Rigley, Yuxuan Liu, Jeffrey Hawke, Richard Shen, Vinh-Dieu
  Lam, and Alex Kendall.
\newblock Learning to drive from simulation without real world labels.
\newblock In {\em ICRA}, 2019.

\bibitem{nuscenes}
Holger Caesar, Varun Bankiti, Alex~H. Lang, Sourabh Vora, Venice~Erin Liong,
  Qiang Xu, Anush Krishnan, Yu Pan, Giancarlo Baldan, and Oscar Beijbom.
\newblock nuscenes: A multimodal dataset for autonomous driving.
\newblock In {\em CVPR}, 2020.

\bibitem{mp3}
Sergio Casas, Abbas Sadat, and Raquel Urtasun.
\newblock {MP3}: A unified model to map, perceive, predict and plan.
\newblock In {\em CVPR}, 2021.

\bibitem{argoverse}
Ming-Fang Chang, John~W Lambert, Patsorn Sangkloy, Jagjeet Singh, Slawomir Bak,
  Andrew Hartnett, De Wang, Peter Carr, Simon Lucey, Deva Ramanan, and James
  Hays.
\newblock Argoverse: 3d tracking and forecasting with rich maps.
\newblock In {\em CVPR}, 2019.

\bibitem{carla}
Alexey Dosovitskiy, German Ros, Felipe Codevilla, Antonio Lopez, and Vladlen
  Koltun.
\newblock {CARLA}: An open urban driving simulator.
\newblock {\em CoRR}, 2017.

\bibitem{waymo}
Scott Ettinger, Shuyang Cheng, and Benjamin~Caine et al.
\newblock Large scale interactive motion forecasting for autonomous driving:
  The {Waymo Open Motion Dataset}.
\newblock {\em arXiv preprint arXiv:2104.10133}, 2021.

\bibitem{kitti}
Andreas Geiger, Philip Lenz, Christoph Stiller, and Raquel Urtasun.
\newblock Vision meets robotics: The {KITTI} dataset.
\newblock {\em IJRR}, 32(11):1231--1237, 2013.

\bibitem{nuscenespkl}
Yiluan Guo, Holger Caesar, Oscar Beijbom, Jonah Philion, and Sanja Fidler.
\newblock The efficacy of neural planning metrics: A meta-analysis of {PKL} on
  nuscenes.
\newblock In {\em IROS Workshop on Benchmarking Progress in Autonomous
  Driving}, 2020.

\bibitem{lyft}
John Houston, Guido Zuidhof, and Luca~Bergamini et al.
\newblock One thousand and one hours: Self-driving motion prediction dataset.
\newblock {\em arXiv preprint arXiv:2006.14480}, 2020.

\bibitem{pointpillars}
Alex~H. Lang, Sourabh Vora, Holger Caesar, Lubing Zhou, Jiong Yang, and Oscar
  Beijbom.
\newblock Pointpillars: Fast encoders for object detection from point clouds.
\newblock In {\em CVPR}, 2019.

\bibitem{cil_cam_lidar_fusion}
Eraqi~Hesham M., Mohamed~N. Moustafa, and Jens Honer.
\newblock Conditional imitation learning driving considering camera and lidar
  fusion.
\newblock In {\em NeurIPS}, 2020.

\bibitem{simulation_reinforcement_learning}
Blazej Osinski, Adam Jakubowski, Pawel Ziecina, Piotr Milos, Christopher
  Galias, Silviu Homoceanu, and Henryk Michalewski.
\newblock Simulation-based reinforcement learning for real-world autonomous
  driving.
\newblock In {\em ICRA}, 2020.

\bibitem{pkl}
Jonah Philion, Amlan Kar, and Sanja Fidler.
\newblock Learning to evaluate perception models using planner-centric metrics.
\newblock In {\em CVPR}, 2020.

\bibitem{transfuser}
Aditya Prakash, Kashyap Chitta, and Andreas Geiger.
\newblock Multi-modal fusion transformer for end-to-end autonomous driving.
\newblock In {\em IEEE Conference on Computer Vision and Pattern Recognition
  (CVPR)}, 2021.

\bibitem{waymo_offboard}
Charles~R. Qi, Yin Zhou, Mahyar Najibi, Pei Sun, Khoa Vo, Boyang Deng, and
  Dragomir Anguelov.
\newblock Offboard 3d object detection from point cloud sequences.
\newblock {\em arXiv preprint arXiv:2103.05073}, 2021.

\bibitem{joint_behavior_trajectory_planning}
Abbas Sadat, Mengye Ren, Andrei Pokrovsky, Yen{-}Chen Lin, Ersin Yumer, and
  Raquel Urtasun.
\newblock Jointly learnable behavior and trajectory planning for self-driving
  vehicles.
\newblock In {\em IROS}, 2019.

\bibitem{airsim2017fsr}
Shital Shah, Debadeepta Dey, Chris Lovett, and Ashish Kapoor.
\newblock {AirSim}: High-fidelity visual and physical simulation for autonomous
  vehicles.
\newblock In {\em Field and Service Robotics}, 2017.

\bibitem{cil_e2e_mmsfs}
Ibrahim Sobh, Loay Amin, Sherif Abdelkarim, Khaled Elmadawy, Mahmoud Saeed,
  Omar Abdeltawab, Mostafa Gamal, and Ahmad~El Sallab.
\newblock End-to-end multi-modal sensors fusion system for urban automated
  driving.
\newblock In {\em NeurIPS}, 2018.

\bibitem{track_with_permanence}
Pavel Tokmakov, Jie Li, Wolfram Burgard, and Adrien Gaidon.
\newblock Learning to track with object permanence.
\newblock {\em arXiv preprint arXiv:2103.14258}, 2021.

\bibitem{cil_multimodal_e2e}
Yi Xiao, Felipe Codevilla, Akhil Gurram, Onay Urfalioglu, and Antonio~M.
  L{\'o}pez.
\newblock Multimodal end-to-end autonomous driving.
\newblock {\em arXiv preprint arXiv:1906.03199}, 2019.

\bibitem{centerpoint}
Tianwei Yin, Xingyi Zhou, and Philipp Kr{\"{a}}henb{\"{u}}hl.
\newblock Center-based 3d object detection and tracking.
\newblock {\em arXiv preprint arXiv:2006.11275}, 2020.

\bibitem{e2e_inmp}
Wenyuan Zeng, Wenjie Luo, Simon Suo, Abbas Sadat, Bin Yang, Sergio Casas, and
  Raquel Urtasun.
\newblock End-to-end interpretable neural motion planner.
\newblock In {\em CVPR}, 2021.

\end{thebibliography}
}

\end{document}